\pdfoutput=1
\documentclass[10pt,twocolumn,letterpaper]{article}

\usepackage{times}
\usepackage{epsfig}
\usepackage{graphicx}
\usepackage{amsmath}
\usepackage{amssymb}
\usepackage{caption}
\usepackage{subcaption}
\usepackage{multirow}
\usepackage{float}

\usepackage[pagebackref=true,breaklinks=true,colorlinks,bookmarks=false]{hyperref}

\newcommand{\eg}{e.g.\ }

\newcommand{\etal}{\textit{et al.\ }}

\begin{document}

\title{3D Human Shape Style Transfer}

\author{ Jo\~ao Regateiro\footnote{Univ. Grenoble Alpes, Inria, CNRS, Grenoble INP (Institute of Engineering Univ. Grenoble Alpes), LJK, 38000 Grenoble, France}\\
{\tt\small j.regateiro@inria.fr}
\and
Edmond Boyer$^*$\\
{\tt\small edmond.boyer@inria.fr}
}

\maketitle

\begin{abstract}
    We consider the problem of modifying/replacing the shape style of a real moving character with those of an arbitrary static real source character. 
    Traditional solutions follow a pose transfer strategy, from the moving character to the source character shape, that relies on skeletal pose parametrization. 
    In this paper, we explore an alternative approach that transfers the source shape style onto the moving character. 
    The expected benefit is to avoid the inherently difficult pose to shape conversion required with skeletal parametrization applied on real characters. 
    To this purpose, we consider image style transfer techniques and investigate how to adapt them to 3D human shapes. 
    Adaptive Instance Normalisation (AdaIN) and SPADE architectures have been demonstrated to efficiently and accurately transfer the style of an image onto another while preserving the original image structure.
    Where AdaIN contributes with a module to perform style transfer through the statistics of the subjects and SPADE contribute with a residual block architecture to refine the quality of the style transfer.
    We demonstrate that these approaches are extendable to the 3D shape domain by proposing a convolutional neural network that applies the same principle of preserving the shape structure (shape pose) while transferring the style of a new subject shape.
    The generated results are supervised through a discriminator module to evaluate the realism of the shape, whilst enforcing the decoder to synthesise plausible shapes and improve the style transfer for unseen subjects.
    Our experiments demonstrate an average of $\approx 56\%$ qualitative and quantitative improvements over the baseline in shape transfer through optimization-based and learning-based methods.
\end{abstract}

\begin{figure}[H]
\begin{center}
    \includegraphics[width=0.65\linewidth]{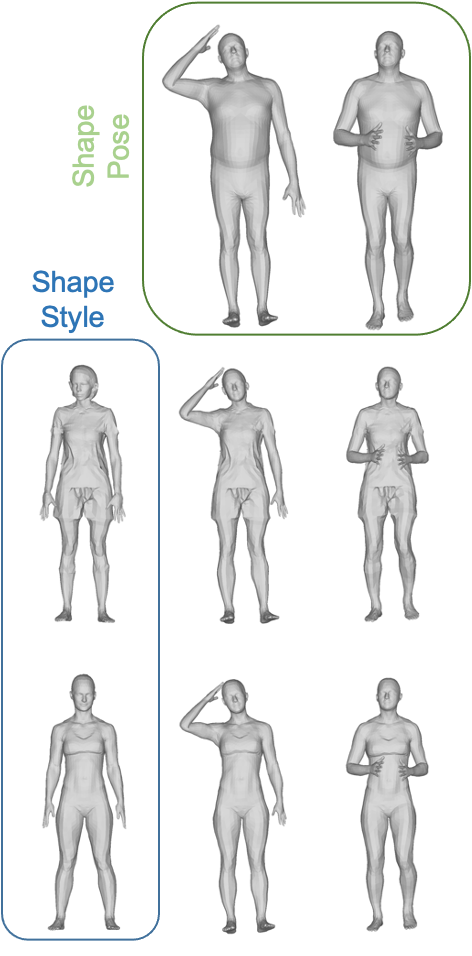}
\end{center}
   \caption{Shape style transfer from two unseen realistic shapes (left-most shape) onto two posed shapes (top shapes). The results demonstrate the ability to transfer realistic unseen shapes.}
\label{fig:highlight}
\end{figure}
\section{Introduction}
Motion retargeting is the process of transferring motion between digital characters. 
This is largely used to create animations based on information captured on real characters and enrich therefore creative applications and also expand existing \cite{Bogo:CVPR:2014,dfaust:CVPR:2017,vonMarcard2018,AMASS:ICCV:2019} datasets.  
Originally based on motion capture data and skeletal parametrization \cite{Lipman2005,Baran2007,Kavan2008} motion retargeting has been extended to full shape information with the progress of computer vision solutions and performance capture systems \cite{Sumner2004,Baran2009,Boukhayma2017}.
While enabling motion retargeting with fully captured shape and motion information such an extension faces an inherent difficulty with traditional retargeting methods based on skeletal parametrization. 
Despite its huge success, this strategy is prone to artefacts and unrealistic surface deformation, such as intra-body intersections, as a consequence of differences in the human body anatomy, such as subtle variations in joint angels between similar poses.
Alternative optimization-based techniques directly deform the surface of the character by transferring the surface from another.
Surface-based motion transfer \cite{Sumner2004,Boukhayma2017,Basset2019} has the objective of disentangling the pose from a character shape, which allows to interchange poses between distinct characters, assuming that pose and shape are not correlated \cite{Anguelov2005}.
This assumption in practice seems incorrect \cite{Basset2019}, as a consequence of the body morphology, which means that identical poses change with the body shape.
Hence, modelling human pose independently of the shape remains an open research topic.

In this paper, we propose to explore image style transfer techniques to the problem of transferring the shape of a source identity to a target character in a given pose, avoiding the problem associated with pose generalization with learning-based methods \cite{Wang_2020_CVPR}.
To this end, we proposed a neural network to transfer the shape style of a character identity onto another character in a given pose, illustrated in Figure \ref{fig:highlight}.
We employ a convolutional neural network to learn such transfer between characters with a common template mesh.
In contrast to the work of Wang \etal \cite{Wang_2020_CVPR}, our method can handle large motion sequences and preserve body contacts without surface distortion.
The proposed neural network learns how to transfer the shape style by aligning the shape and pose statistics between characters.
This idea was inspired by the success of 2D image style transfer of SPADE \cite{Park_2019_CVPR}, which employs a set of stacked convolutional layers followed by an adaptive residual block that guides the style of an image to be present in the final result.
This work transfers this methodology to the 3D domain and contributes with a discriminator network that evaluates the output shape while forcing the decoder to produce realistic body shapes.
This is achieved through an adversarial strategy, which shows an average of $\approx 56\%$ improvement on the shape quality compared to state-of-the-art methods \cite{Wang_2020_CVPR,keyang20unsupervised,Basset2019}

The proposed approach focuses on learning multiple characters' shapes using learning-based methods to facilitate reusing or augment publicly available datasets.
The primary novel contributions of this paper are:
\begin{itemize}
	\item A convolutional neural network to learn how to transfer a character's shape style onto another posed character mesh.
	\item Efficient losses on the character shape and pose to guarantee accurate reconstruction and preserve local and global vertex relationships.
	\item Adversarial strategy on the character shape to enforce realistic synthesis and improve the generator shape style transfer for unseen shapes.
	%
\end{itemize}
\section{Related Work}
Transferring motion between digital characters have long been studied and new approaches continue to emerge.
This subject can be divided into two main groups: skeletal motion transfer or surface motion transfer.
Motion transfer through skeletal parameterization deforms a surface mesh given different joint configurations \eg the transfer is dependent on the relation between the skeletal structure and surface mesh.
Contrarily, surface motion transfer is only guided through affine transformation on each primitive on the surface mesh.
This work follows the direction of surface motion transfer to avoid the issues generated from skeletal structures.
For example, unrealistic surface deformations are caused by blending techniques.
Lastly, skeletons require a great effort to be estimated accurately from real character's content.
In this section, we mainly focus on works most relevant for the problem of transferring the shape between subjects.
\subsection{Skeletal Motion Transfer}
Motion transfer using skeletal parameterization, \eg modifying joint angles to match the desired pose target, aims to generate motion sequences for a novel character. 
Skeleton parameterization allows the transfer of motion between rigged meshes, \eg a skeletal structure is attached to the mesh surface through skinning weights \cite{Yu2004,Lipman2005} and deformed using blending techniques \cite{Kry2002,Joshi2006,Wang2007,Baran2007,Kavan2008} to match the pose of the shape target, which has been popular for the re-animation of digital characters.
However this approach is limited to pose transfer, does not provide shape style transfer from new identities \cite{Yan2006, Xiao2008, Aberman2020}, and is commonly known to produce unrealistic surface deformations \cite{Baran2007,Kavan2008}.
\subsection{Shape Motion Transfer}
To overcome the limitation of skeletal motion transfer, researchers focused on transferring the shape style onto a target via exploring properties of the shape geometry \cite{Sumner2004,Baran2009,Zhou2010,Rhodin2015,Boukhayma2017,gaovcgan2018}.
The works of Sumner \etal and Baran \etal \cite{Sumner2004,Zhou2010} mostly encode the pose of the source character as the deformation of the surface mesh, and transfer it to the target character, through per-triangle affine transformations assuming correspondence. 
This approach produces artefacts when transferring between significantly different shapes. 
Other works explored semantic deformation transfer between characters' surfaces allowing transfer between very distinct shapes.
These methods usually consider the pose encoded in the shape of a source character, consequently exploiting pose correspondences \cite{Baran2009}.
Others, explore animations \cite{Boukhayma2017} between characters to define semantic correspondences to compute the transfer of new source poses to the corresponding target character.
More recently, Basset \etal \cite{Basset2019} presented a robust optimization-based method to deform a source character shape and pose using shape similarity, volume preservation and body parts collision as an energy function.
Their method explores shape style transfer instead of pose transfer, hence being the baseline method for comparison.
%
%
%
\subsubsection{Learning-based Motion Transfer}
More recently, neural networks demonstrated the ability to learn deformation transfer from data \cite{gaovcgan2018,keyang20unsupervised}, where a network is trained on several examples of source and target poses.
It is also possible to learn mappings between semantically different poses of humans and animals to interactively control animation generation \cite{Rhodin2015}. 
These methods neither require skeletons nor point-to-point correspondences between source and target. 
However, heavy pre-processing needs to be performed for every pair of source and target characters.
Learning-based methods are required to be re-trained to handle novel identities or fail to generalize to unseen examples \cite{keyang20unsupervised}, whereas we propose to learn multiple shapes with a base template that allows generalization to new identities. 

Most learning-based approaches focus on transferring pose deformation between characters \cite{Wang_2020_CVPR,keyang20unsupervised}, which can lead to unrealistic shape deformation when given extreme unseen poses, therefore is limited to a minimum range of motions.
Wang \etal \cite{Wang_2020_CVPR} proposed to reuse concepts from 2D image style transfer to learn a spatial adaptive network that is invariant to vertex order, although their method strongly suffers from stretching artefacts when the identity shape has body contacts or limbs in proximity.
Therefore, not being suitable to shape transfer for long sequences of realistic motion capture data.
Whereas in the proposed work we tackle the inverse problem, where the posed shape deforms to adopt the style of the identity shape, avoiding the problem encountered when performing large deformation between distinct poses.
Further, our method is capable to robustly transfer unseen shape styles to long mesh sequences without suffering from stretching or unrealistic artefacts.
Hence, being preferable over the state-of-the-art for shape style transfer and motion capture data augmentation.
\section{Deep Shape Style Transfer}
\label{sec:deepsst}
\begin{figure}[t]
\begin{center}
    \includegraphics[width=1.0\linewidth]{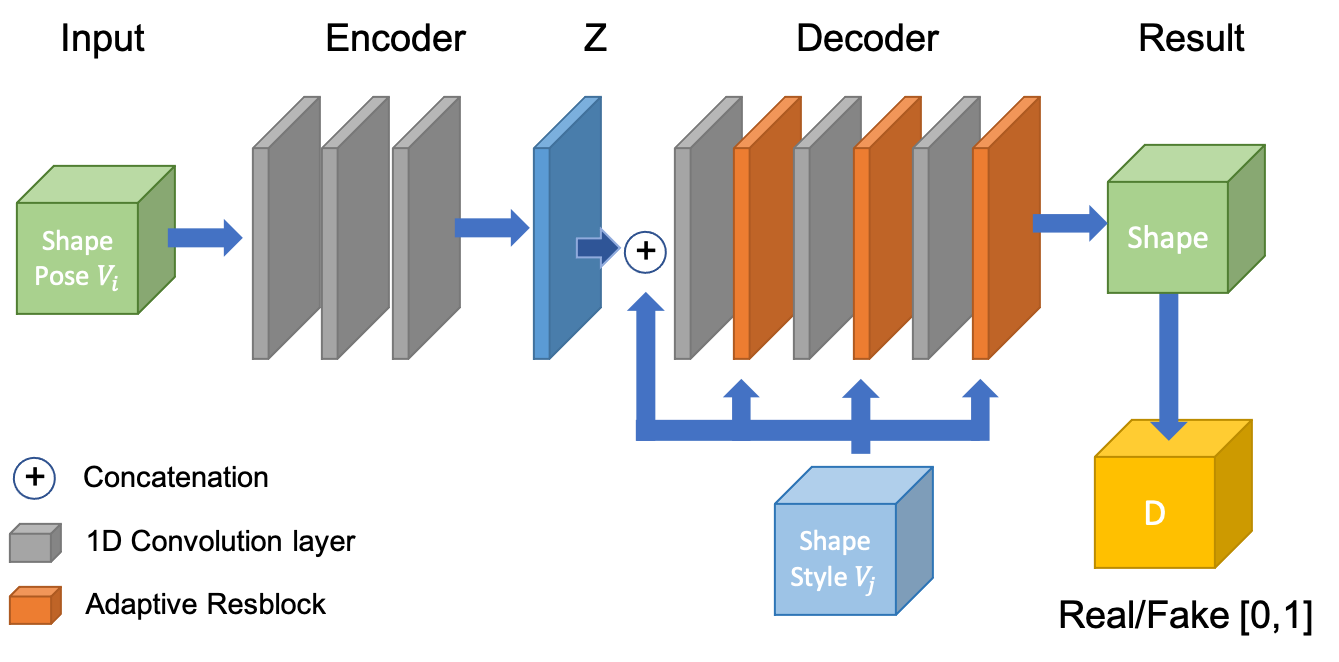}
\end{center}
   \caption{Shape Style Transfer Network receives as input a posed shape and an identity shape style. The encoder extracts relevant pose features. The decoder processes shape pose features and an identity style to align the shape morphology with the posed shape. The output is a deformed posed shape to match the shape style of a novel identity.}
\label{fig:framework}
\end{figure}
In this section, we present the proposed neural network for shape style transfer between a posed shape and an identity shape.
As illustrated in Figure \ref{fig:framework} our network first extracts shape pose features from the input mesh with the desired pose.
This is achieved with an encoder presented in Section \ref{sec:autoencoder}.
Given the pose feature and a shape style, a decoder (Section \ref{sec:decoder}) predicts the input mesh deformation that will transfer the desired identity.
Such deformation can be unrealistic for unseen identities and a discriminator  (Section \ref{sec:discriminator}) completes the architecture with the aim of improving quality.
Our experiments (Section \ref{sec:resultsevaluations}) validate this architecture and demonstrate improvement over traditional optimisation-based shape transfer \cite{Basset2019} as well as over the state-of-the-art learning-based methods \cite{Wang_2020_CVPR,keyang20unsupervised} for shape transfer.
\subsection{Input Data}
In the context of this work, we exploit temporally consistent 3D meshes $\{M^{i} \}_{i=1}^{N}$ of the same connectivity, where $N$ is a collection of meshes with different poses and identity.
A 3D mesh is represented as $M = \{V,\theta\}$, where $V=[v^1, v^2, ..., v^{N_v]}$ is a collection of $N_v$ vertices, $\theta$ is the mesh pose.
The input of the encoder presented in Section \ref{sec:autoencoder} consist of a single 3D mesh $M^{i}=\{V^{i},\theta^{i}\}$, where a collection of stacked convolutional layers process the mesh as a point-cloud to learn spatial pose features.
The input of the decoder consist of a source latent feature $z^{i}$ and a target identity 3D mesh $M^{j}=\{V^{j},\theta^{j}\}$.
Network architecture is illustrated in Table \ref{tab:encoder}.
\subsection{Encoder Shape Features Extraction}
\label{sec:autoencoder}
\begin{table}[b]
	\centering
	\caption{Shape style transfer network architecture.}
	\label{tab:encoder}
	\resizebox{1.0\columnwidth}{!}{%
	\begin{tabular}{cccc}
		\hline
		Encoder Layers&Input Size&Output Size&Activation\\
		\hline
		1x1 Convolution & $3\times x$&$64 \times x $& ReLU\\
		1x1 Convolution & $64 \times x$&$128 \times x$& ReLU\\
		1x1 Convolution & $128 \times x$&$1024 \times x$& ReLU\\
		\hline
	\end{tabular}}
	\resizebox{0.9\columnwidth}{!}{%
	\begin{tabular}{cccc}
		\hline
		Decoder Layers&Input Size&Output Size&Activation\\
		\hline
		1x1 Convolution & $1024\times x$&$1024 \times x $& Instance Norm\\
		Adaptive ResBlock & $1024\times x$&$1024 \times x $& -\\
		1x1 Convolution & $1024 \times x$&$512 \times x$& Instance Norm\\
		Adaptive ResBlock & $512\times x$&$512 \times x $& -\\
		1x1 Convolution & $512 \times x$&$256 \times x$& Instance Norm\\
		Adaptive ResBlock & $256\times x$&$256 \times x $& -\\
		1x1 Convolution & $256\times x$&$3 \times x$& Tanh\\
		\hline
		\hline
	\end{tabular}}
\end{table}
This section describes the module that extracts information from a source mesh $M^{i}$ of which the pose needs to be preserved but the identity transformed.
Such an encoder is expected to extract shape features that, when combined with an identity mesh, enable the prediction of an identity deformation.
These features are in principle close to identity free shape features, such as skeleton pose parametrization, without however the constraints brought by an explicit representation.
Inspired by PointNet \cite{Charles2017} the encoder illustrated in Figure \ref{fig:framework} has the objective to extract shape features $z^{i}$ from the given input shape.
The adaptation from PointNet \cite{Charles2017} allows the encoder to directly focus on the spatial proprieties of each vertex $\{v^{i} \}_{i=1}^{N_v}$.
In the context of this work the input shape $M^{i}=\{V^{i},\theta^{i}\}$ is represented as a collection of vertices $\{V^{i} \}_{i=1}^{N}$, mostly know as a point-cloud.
Hence, the latent space $z^{i}$ represents spatial features for every vertex $v \in V^{i}$ from the input shape $M^{i}$.
This is an important step, as we wish to preserve the structural features in the latent space and allow the decoder to transfer the identity shape style while preserving the general pose structure.
\subsection{Shape Style Transfer Decoder}
\label{sec:decoder}
The decoder has the objective to generate realistic shapes that have the pose of the input data and the shape style of the identity input shape.
The problem is to deform the vertices $V^{i}$ of the source mesh $M^{i}=\{V^{i},\theta^{i}\}$ to match the shape style $V^{j}$ of the identity mesh $M^{j}=\{V^{j},\theta^{j}\}$, while preserving the pose $\theta^{i}$ of the source mesh.
This problem shares strong similarities with image style transfer in the image domain \cite{Xu2017,Park_2019_CVPR,Karras2019}.
The approaches that highly motivated this work are adaptive instance style transfer \cite{Xu2017}, the style-based generator \cite{Karras2019} and residual network blocks for style transfer \cite{Park_2019_CVPR}.
The following sections describe in more detail each block of the shape style transfer decoder and the training details.
\subsubsection{Instance Normalization}
\begin{figure}[t]
\begin{center}
    \includegraphics[width=0.75\linewidth]{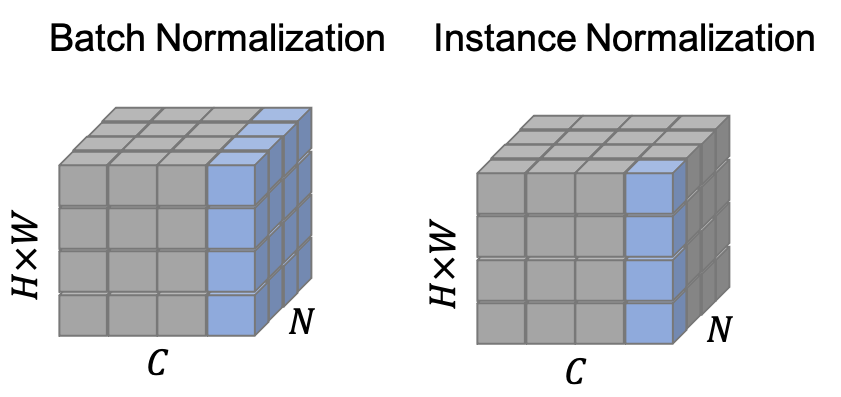}
\end{center}
   \caption{This figure shows on the left Batch Normalization (Batchnorm) and on the right Instance Normalization (IN) visual illustration.}
\label{fig:bnin}
\end{figure}
The problem is to guarantee that high-frequency information from a style shape is preserved and transferred, without blurring the data with statistics from other samples in a batch.
The Batch Normalization (Batchnorm) layer is commonly known for computing statistics across a batch \eg more than one sample, which can lead to the loss of high-frequency information present in the data, hence the choice to use Instance Normalization (IN).
IN plays an important role in our framework, which is to gather statistics of individual samples.
In our framework, we want to consider the information of a single mesh and how the identity style can affect the posed mesh.
To understand how IN works we directly compare it with commonly used Batchnorm.
Figure \ref{fig:bnin} shows an example of the behaviour of both Batchnorm and IN layers.
Normalization layers usually take the output $X$ of a convolutional layer and transform it to have mean zero and unit standard deviation.
Equation \ref{eq:normalization} is standard across both normalization techniques, where $X$ represents the input data, $\mu$ is the mean of $X$, $\sigma$ is the standard deviation of $X$ and $\epsilon$ is a constant value added for numerical stability.
The major difference between the two methods is the way they process the information, for example, Batchnorm looks across the height ($H$) and width ($W$) of the data, illustrated in Figure \ref{fig:bnin} with light blue quads.
For each channel, $C$ Batchnorm considers $N$ samples, representative of a batch, to compute the mean and standard deviation of the data.
This is then processed per individual channel for all batches, hence the name batch normalization.

On the other hand IN only considers one instance of the data instead of a batch (Figure \ref{fig:bnin}).
In this situation, we want to gather statistics of a single channel $C$ at one instance of the batch $N$.
Looking at the Equation \ref{eq:normalization} the main difference between the two methods would be the representation of the data $X$, which would be modified to $X^{i}$ accordingly to reflect a single sample $i$ of the data.
\begin{equation}
\label{eq:normalization}
Y = \dfrac{X - \mu(X)}{\sqrt{\sigma(X) + \epsilon}}
\end{equation}
\subsubsection{Adaptive Residual Block}
\begin{figure}[H]
\begin{center}
    \includegraphics[width=.75\linewidth]{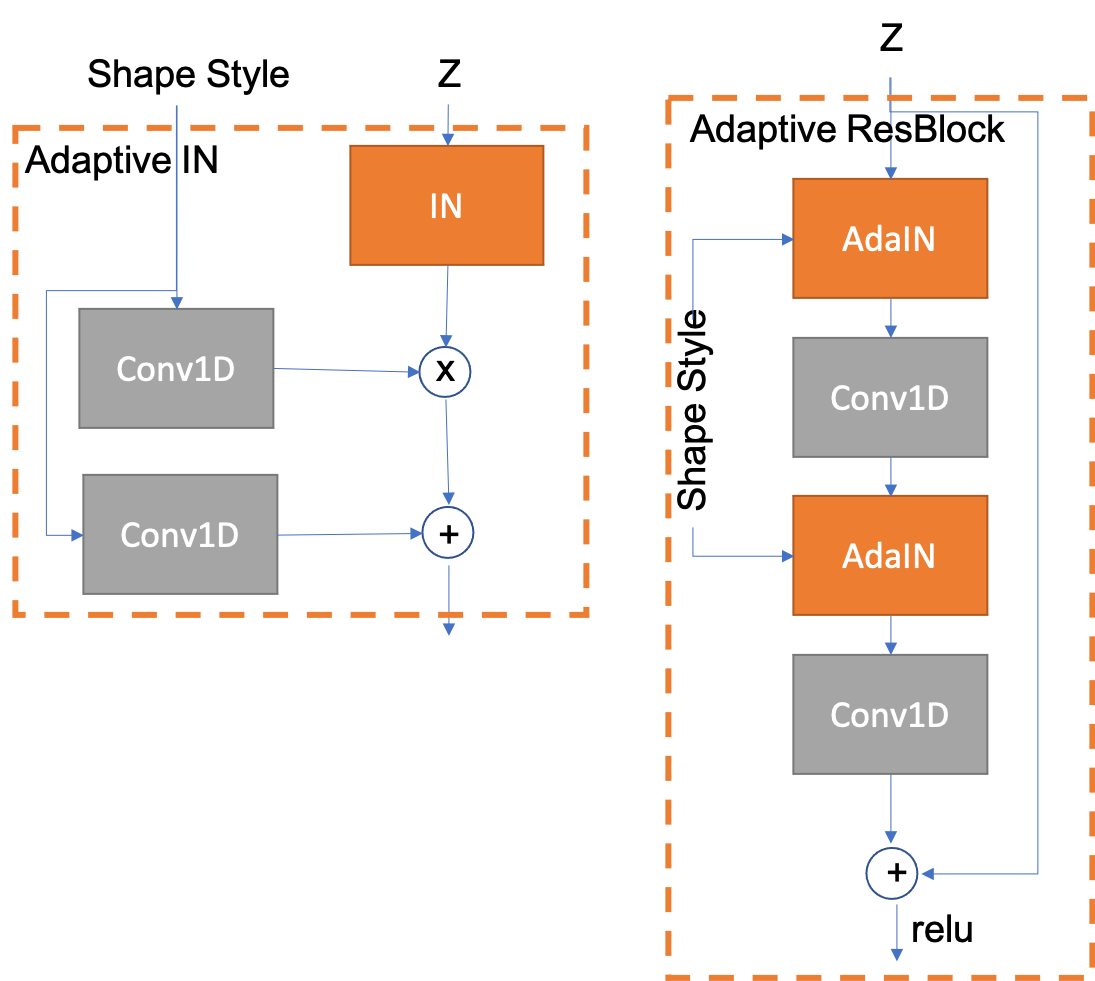}
\end{center}
   \caption{This figure shows on the left Adaptive Instance Normalization \cite{Xu2017} and on the right adaptive residual block \cite{Park_2019_CVPR} architecture.}
\label{fig:adaresblock}
\end{figure}
Residual block architecture has been shown to preserve semantic information present in the image domain \cite{Park_2019_CVPR}, however only having a normalization layer, such as the IN, tend to wash away semantic information when applied to uniform or flat segmentation masks.
In the work of Park \etal \cite{Park_2019_CVPR} this is clearly illustrated with a simple example, where a module first applies convolution and then normalization.
This test demonstrates that using normalization over uniform values loses the semantic information, while the proposed adaptive residual block can better preserve semantic information by using an adaptive instance normalization architecture.
Given the efficacy of the method over the image domain, we demonstrate how to transfer the same concept over to the 3D shape domain to preserve the shape style.
The adaptive residual block is illustrated in Figure \ref{fig:adaresblock}.
\subsubsection{Adaptive Instance Normalization}
Adaptive Instance Normalization (AdaIN) \cite{Xu2017} is commonly known as conditional instance normalization, because it makes use of the Instance Normalization layer and guides the results given an external factor, which we described as style, illustrated in Figure \ref{fig:adaresblock}.
AdaIN was developed as an extension of IN to allow the representation of arbitrarily styles by using adaptive affine transformations.
Given and input $X^{i}$ and shape style $Y$ the goal is to align the channel-wise mean and variance of $X^{i}$ to match those of $Y$, as in Equation \ref{eq:adain}.
Unlike other normalization techniques, AdaIN has no learnable affine parameters.
Instead, it adaptively computes the affine parameters from
the style input $Y$ \cite{Xu2017}.
AdaIN performs style transfer in the feature space by transferring feature statistics, specifically the channel-wise mean and variance.
\begin{equation}
\label{eq:adain}
AdaIN(X^{i} ,Y) = \sigma(Y) \dfrac{X^{i} - \mu(X^{i})}{\sqrt{\sigma(X^{i}) + \epsilon}} + \mu(Y)
\end{equation}
\subsection{Shape Features Discriminator}
\label{sec:discriminator}
The network can produce unrealistic shapes for unseen samples, hence we follow an adversarial strategy as in GAN architectures to improve the quality of the generated shapes.
A GAN architecture is relatively simple and widely used for generative models.
It is composed of a generator and a discriminator network, where these two work together to improve each other predictions.
The discriminator is trained on real and fake data, hence being responsible to distinguish between real and generated data.
On the other hand, the generator is trained via the discriminator predictions.

In the context of this work, the discriminator has the objective to classify the shape as real or fake, synthesised by the decoder.
This enforces the network to synthesise shapes that are in accordance with the training data and generalises better to unseen identities.
The discriminator tends to avoid aberrant results which do not yield a high percentage of improvement on average but appear beneficial in practice.
This strategy is validated experimentally with $\approx 5\%$ of improvement as shown in the evaluation Section \ref{sec:resultsevaluations}.
\section{Training Details}
The network is trained with an end-to-end fashion, where the encoder, decoder and discriminator models are learned simultaneously, as illustrated in Figure \ref{fig:framework}.
Equation \ref{eq:totalloss} defines the loss function minimized by the network to solve for the task of shape style transfer.
Equation \ref{eq:dtotalloss} defines the adversarial loss function that simultaneously trains the discriminator, encoder and decoder models to synthesise realistic shapes.
\begin{equation}
\label{eq:totalloss}
\begin{split}
\mathcal{L}_{g} = \vartheta_{adver} \mathcal{L}_{adver} + \vartheta_{rec} \mathcal{L}_{rec} + \\
\vartheta_{edge} \mathcal{L}_{edge} + \vartheta_{dist} \mathcal{L}_{dist}
\end{split}
\end{equation}
\begin{equation}
\label{eq:dtotalloss}
\begin{split}
\mathcal{L}_{d} = \mathcal{L}_{minimax}
\end{split}
\end{equation}
Where, $\vartheta_{adver} = 0.1$, $\vartheta_{rec} = 2.0$, $\vartheta_{edge} = 0.5$ and $\vartheta_{dist} = 2.0$ are scalar weights to balance the contribution of each $\mathcal{L}$ component.
%

The component $\mathcal{L}_{rec}$ guarantees the result to be real by approximating the synthesised sample to its ground truth as close as possible.
This permits preservation of the global axis location of the generated sample and difficult the discriminator task to distinguish a real from a fake shape.
Equation \ref{eq:reconloss} represents the L2 distance between the generated sample and the ground truth identity style.
\begin{equation}
\label{eq:reconloss}
\mathcal{L}_{rec} = \sum || V^{i} - \tilde{V^{i}}||^2 
\end{equation}
%

The component $\mathcal{L}_{edge}$ enforces smoothness over the mesh surface by penalizing longer edges, which has been demonstrated efficient for such regularization \cite{Groueix2018}.
This regularization enforces the generated mesh surface $\tilde V^{i}$ to be similar to the shape style $V^{j}$ independent of the pose parameter.
Hence, being a good candidate to enforce the transfer of shape style.
Equation \ref{eq:edgeloss} defines the edge length regularization, which enforces edges to keep the same length between the identity $V^{j}$  and the generated $\tilde V^{i}$ shape.
Where $\mathcal{N}(\tilde v^{i})$ is the one ring neighbor of vertex $\tilde v^{i} \in \tilde V$.
\begin{equation}
\label{eq:edgeloss}
\mathcal{L}_{edge} = \sum_{i \in V} \sum_{ j \in \mathcal{N}(\tilde v^{i})} || \tilde v^{i} - v^{j}||^2 
\end{equation}
%

The component $\mathcal{L}_{dist}$ focus on the shape pose and distance between body parts by looking at the dense distance matrix between the mesh vertices.
The objective is to preserve the global structure of the shape instead of the surface, such as the human pose and the relation between body contact and relative distances between limbs.
This component tends to preserve the shape volume, hence its contribution needs to be adjusted to allow the results to adapt to the new shape identity if the ground truth is not known.
Otherwise, the method should allow the transfer of a new identity while strongly preserving the shape pose.
Equation \ref{eq:distloss} defines the shape pose loss, by comparing the distance matrix $\mathcal{U}^{i,j}(.)$ of the generated $\tilde V^{i}$ and the ground-truth shape $\mathcal{V}^{i}$, where $n$ is the number of vertices in the upper triangle of the matrix excluding the diagonal.
The distance matrix $\mathcal{U}^{i,j}(.)$ is computed by using the euclidean distance between mesh vertices, \eg every vertex on the mesh will have the euclidean distance to all other vertices and vice versa.
\begin{equation}
\label{eq:distloss}
\mathcal{L}_{dist} = \dfrac{1}{n} \sum|| \mathcal{U}^{i,j}(\mathcal{V}^{i}) - \mathcal{U}^{i,j}(\tilde V^{i})||^2 
\end{equation}
The discriminator module $\mathcal{L}_{d}$ is trained to maximize the probability of predicting the correct label of both training examples and the decoder synthesis \cite{Goodfellow2020}.
This is implemented as a binary classification problem with binary labels $0$ and $1$ for generated and real shapes respectively.
Equation \ref{eq:minmax} refers to the simultaneous optimization of the discriminator and decoder models.
\begin{equation}
\label{eq:minmax}
\begin{split}
\mathcal{L}_{minmax} = \min_G \max_D \mathbb{E}_{M \sim p(V)}  [\log(D(V)] + \\
\mathbb{E}_{z \sim p(z)}  [\log(1 - D(G(z)))]
\end{split}
\end{equation}
The min and max refer to the minimization of the decoder loss and the maximization of the discriminator's loss.
The discriminator seeks to maximize the average of the log probability of real samples and the log of the inverse probability for the fake samples, as follows,
\begin{equation}
\label{eq:discloss}
\begin{split}
\mathcal{L}_{disc} = \max_D \mathbb{E}_{M \sim p(V)}  [\log(D(V)] + \\
\mathbb{E}_{z \sim p(z)}  [\log(1 - D(G(z)))]
\end{split}
\end{equation}
The decoder on the other hand minimizes the log of the inverse probability predicted by the discriminator for fake samples, as follows,
\begin{equation}
\label{eq:adver}
\begin{split}
\mathcal{L}_{adver} = \min_G \mathbb{E}_{z \sim p(z)}  [\log(1 - D(G(z)))]
\end{split}
\end{equation}
This encourages the decoder to synthesise shapes that have a low probability of not being real.
\begin{table*}[!h]
	\centering
	\caption[Average shape distance error evaluation for all datasets]{ \label{tab:ErrorEvaluations}
		Comparison of the proposed network, NPT \cite{Wang_2020_CVPR}, USPD \cite{keyang20unsupervised} and CPST \cite{Basset2019}. The values represent the Hausdorff (HDFF) average and root mean squared (RMSE) error in meters (m) across all motion sequences for different characters' shapes. The best results are illustrated in bold.}
\resizebox{1.0\linewidth}{!}{%
		\begin{tabular}{*{18}{c}}  
			\hline
			\multicolumn{1}{c}{} &  
			\multicolumn{2}{c}{{Proposed (with D)}} &
			\multicolumn{2}{c}{{Proposed (without D)}} &
			\multicolumn{2}{c}{{NPT maxpool \cite{Wang_2020_CVPR}}} &
			\multicolumn{2}{c}{{NPT \cite{Wang_2020_CVPR}}} &
			\multicolumn{2}{c}{{USPD \cite{keyang20unsupervised}}} &
			\multicolumn{2}{c}{{CPST \cite{Basset2019}}}\\  
			\multicolumn{1}{c}{{Dataset}} & \multicolumn{1}{c}{{HDFF}(m)}  & \multicolumn{1}{c}{{RMSE}(m)} & \multicolumn{1}{c}{{HDFF}(m)} &
			\multicolumn{1}{c}{{RMSE}(m)} & \multicolumn{1}{c}{{HDFF}(m)} &
			\multicolumn{1}{c}{{RMSE}(m)} & \multicolumn{1}{c}{{HDFF}(m)} &
			\multicolumn{1}{c}{{RMSE}(m)} &
			\multicolumn{1}{c}{{HDFF}(m)} &
			\multicolumn{1}{c}{{RMSE}(m)} &
			\multicolumn{1}{c}{{HDFF}(m)} &
			\multicolumn{1}{c}{{RMSE}(m)}\\ \hline 
			Training    & 0.0289 & 0.0075 & \textbf{0.0234} & \textbf{0.0055} & 0.0547 & 0.0155 & 0.0617 & 0.01553 & 0.1769 & 0.0704 & 0.0848 & 0.0368  \\ 
            Validation  & \textbf{0.0284} & \textbf{0.0069} & 0.0305 & 0.0071 & 0.0357 & 0.0092 & 0.0429 & 0.0097 & 0.0528 & 0.0201 & 0.0799 & 0.0328    \\
			\hline
	\end{tabular}
}
\end{table*}
\section{Evaluation}
\label{sec:resultsevaluations}
This section presents results and evaluation for the proposed shape style transfer network, introduced in Section \ref{sec:deepsst}.
To evaluate the proposed architecture we use AMASS \cite{AMASS:ICCV:2019}, a large database of human motion that combines optical marker-based motion capture and realistic 3D human meshes.
AMASS \cite{AMASS:ICCV:2019} exploits SMPL model \cite{SMPL:2015}, which provides skeletal representation and fully rigged surfaces mesh, allowing to generate distinct body shape while preserving the same motion.
SMPL based datasets are used to evaluate generalization capabilities, such as FAUST, Dynamic FAUST \cite{Bogo:CVPR:2014,dfaust:CVPR:2017} and realistic clothed people (3DPW) \cite{vonMarcard2018}, which are challenging examples.
Figures \ref{fig:highlight}, \ref{fig:test1} , \ref{fig:test2}, \ref{fig:clothingval2} and \ref{fig:clothingval5}, demonstrate the ability to represent such challenging datasets on realistic and clothed shapes.

A comparison against learning-based and shape transfer optimisation methods \cite{Wang_2020_CVPR, keyang20unsupervised, Basset2019} is shown in Table \ref{tab:ErrorEvaluations} and illustrated in Figure \ref{fig:eval_compare_results}, demonstrating $\approx 56\%$ improvement on the shape quality.
Table \ref{tab:ErrorEvaluations} illustrates an ablation study on the proposed network, to quantitative evaluate the contribution of each module, which shows $\approx 5\%$ of improvement for unseen samples using the proposed model.
\subsection{Datasets}
\label{sec:datasets}
The AMASS dataset \cite{AMASS:ICCV:2019} used for training consists of eighteen different body shapes, each with $21$ different motion sequences.
This dataset uses the SMPL model \cite{SMPL:2015} to represent the body shape of each distinct shape and pose.
SMPL is a linear model based on skinning and blend shapes learned from thousands of human body scans, allowing multiple humans with distinct body shapes to be represented by a single template model.
The $18$ body shapes were randomly generated to create a unique collection of shapes, the poses were generated using motion capture data to allow each shape to have the same motions.
This procedure provides ground-truth data to train and to evaluate the performance of the network, see the supplementary for shape and poses diversity used for training and testing.
\subsection{Quantitative Evaluation}
The proposed network uses Equations \ref{eq:totalloss} and \ref{eq:dtotalloss} as a metric to transfer the shape of an identity character onto a posed mesh.
We compare the shape transfer with ground truth samples acquired from using the AMASS model \cite{AMASS:ICCV:2019} that allows distinct shape synthesis augmented with motion capture data.
Shape style evaluation is performed using the Root Mean Squared Distance (RMSD) defined as $RMSD (A,B) = \sqrt{\frac{\sum^{n}_{i=1}{(B_i - A_i)^2} }{n}} $ between the generated $B$ and ground truth $A$ mesh vertices, which provides an approximation measurement between meshes.
To evaluate the pose representation and distance to shape we use Hausdorff distance defined as $ d_H (A,B) = max \{ \sup_{a \in A} d( a,B ), \sup_{b \in B} d( b,A ) \} $, where $d(a,B)$ and $d(b,A)$  is the distance from a point $a$ to a set $B$ and from a point $b$ to a set $A$, which has been shown to be a good measurement between 3D meshes.
The evaluation contains training and validation data for all characters shape styles, see Table \ref{tab:ErrorEvaluations}.

\begin{figure}[H]
    \begin{center}
        \includegraphics[width=1.0\linewidth]{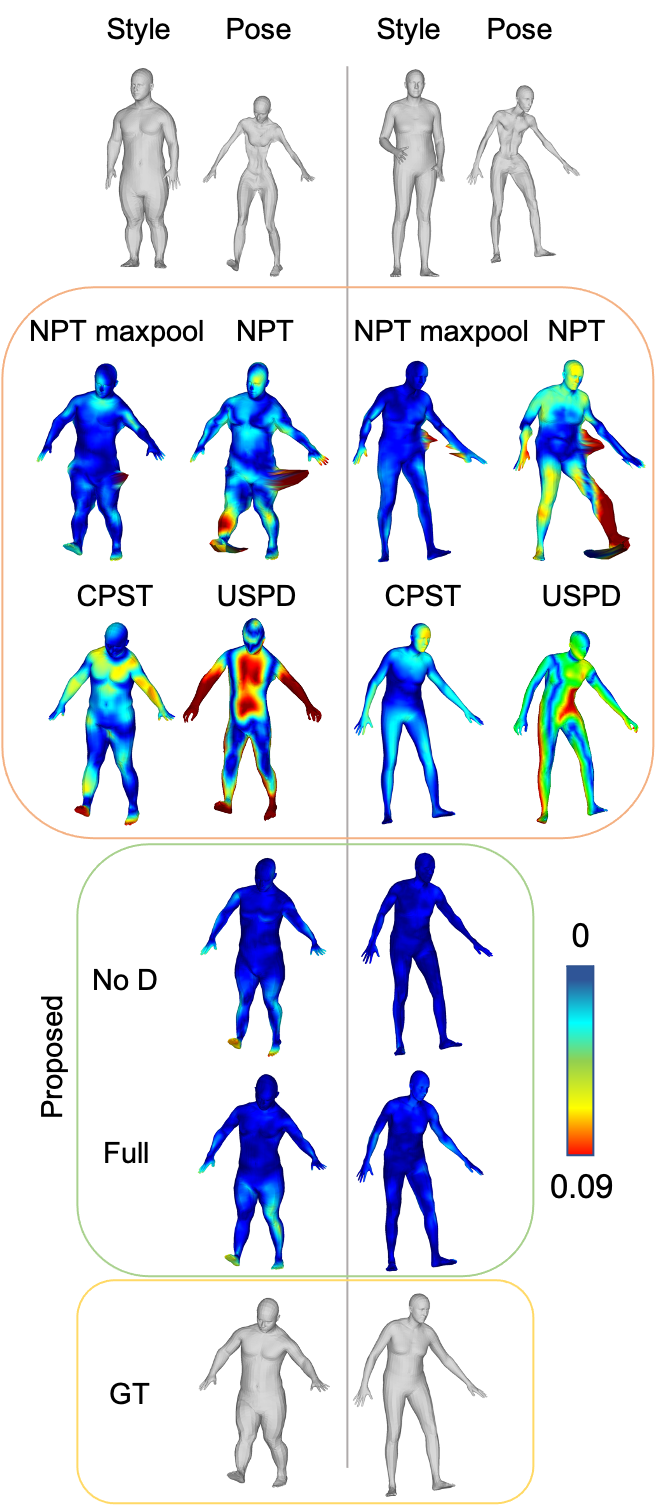}
    \end{center}
       \caption{Shape style transfer comparison on training (right) and validation (left) data with state-of-the-art method NPT \cite{Wang_2020_CVPR}, USPD \cite{keyang20unsupervised} and CPST \cite{Basset2019}.}
    \label{fig:eval_compare_results}
\end{figure}

We have compared our method with and without the discriminator model, demonstrating that the proposed network benefits from having the discriminator model to evaluate the quality of the generated shape and hence the results improve for unseeing shape styles.
We have compared our method with three state-of-the-art methods, Neural Pose Transfer (NPT) \cite{Wang_2020_CVPR}, Unsupervised Shape and Pose Disentanglement (USPD) \cite{keyang20unsupervised} and Contact Preserving Shape Transfer (CPST) \cite{Basset2019}.
Most similar to our approach NPT \cite{Wang_2020_CVPR} is a learning-based method that learns to deform the identity shape to match the posed character.
Whereas, we propose to deform the posed character shape to match the shape style of the identity, avoiding issues of unrealistic mesh deformations and stretching artefacts, illustrated in Figure \ref{fig:eval_compare_results}.
USPD \cite{Wang_2020_CVPR} is a learning-based method that disentangles shape and pose from 3D meshes using auto-encoders architectures.
Unlike the proposed method, USPD \cite{Wang_2020_CVPR} requires many examples of body shapes to allow generalisation to unseen styles, hence not being able to handle large changes in body shape, illustrated in Figure \ref{fig:eval_compare_results}.
In our experiments, Table \ref{tab:ErrorEvaluations}, we demonstrate quantitative and qualitative improvement over learning-based methods, and comparable shape transfer results over the base-line optimisation method CPST \cite{Basset2019}, although the proposed method drastically improves on the performance on shape transfer over the base-line method.
\subsection{Qualitative Evaluation}
\begin{figure}[t]
\begin{center}
    \includegraphics[width=0.90\linewidth]{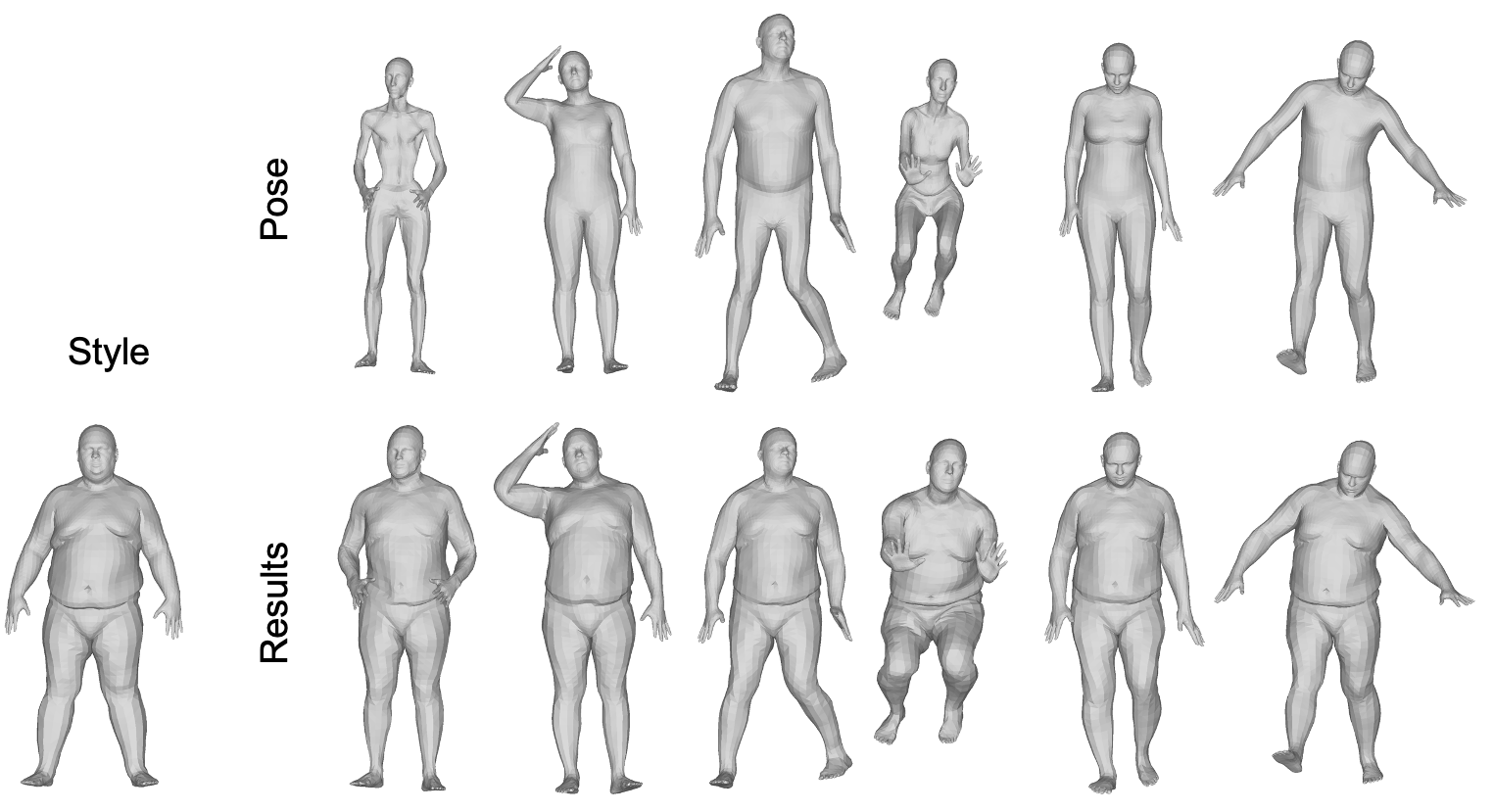}
\end{center}
   \caption{Large body style transferred to different poses with varied body shapes.}
\label{fig:test1}
\end{figure}

\begin{figure}[t]
\begin{center}
    \includegraphics[width=0.90\linewidth]{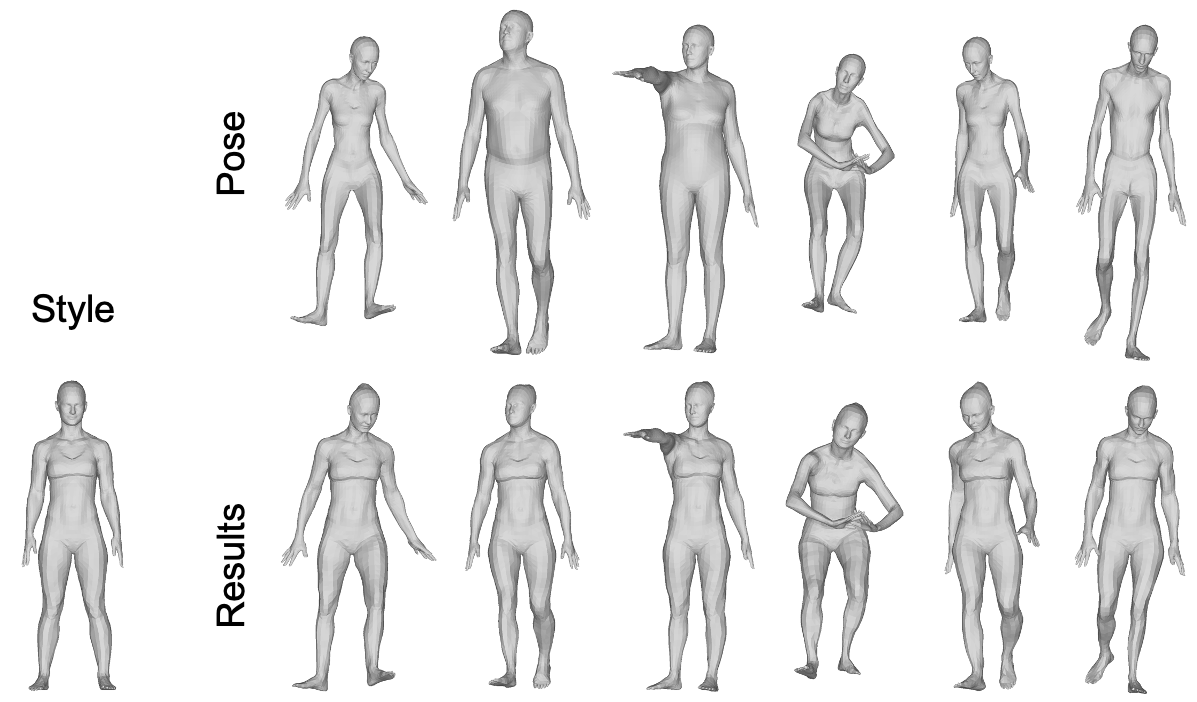}
\end{center}
   \caption{Female body style transferred to different poses with varied body shapes.}
\label{fig:test2}
\end{figure}

\begin{figure}[t]
\begin{center}
    \includegraphics[width=0.80\linewidth]{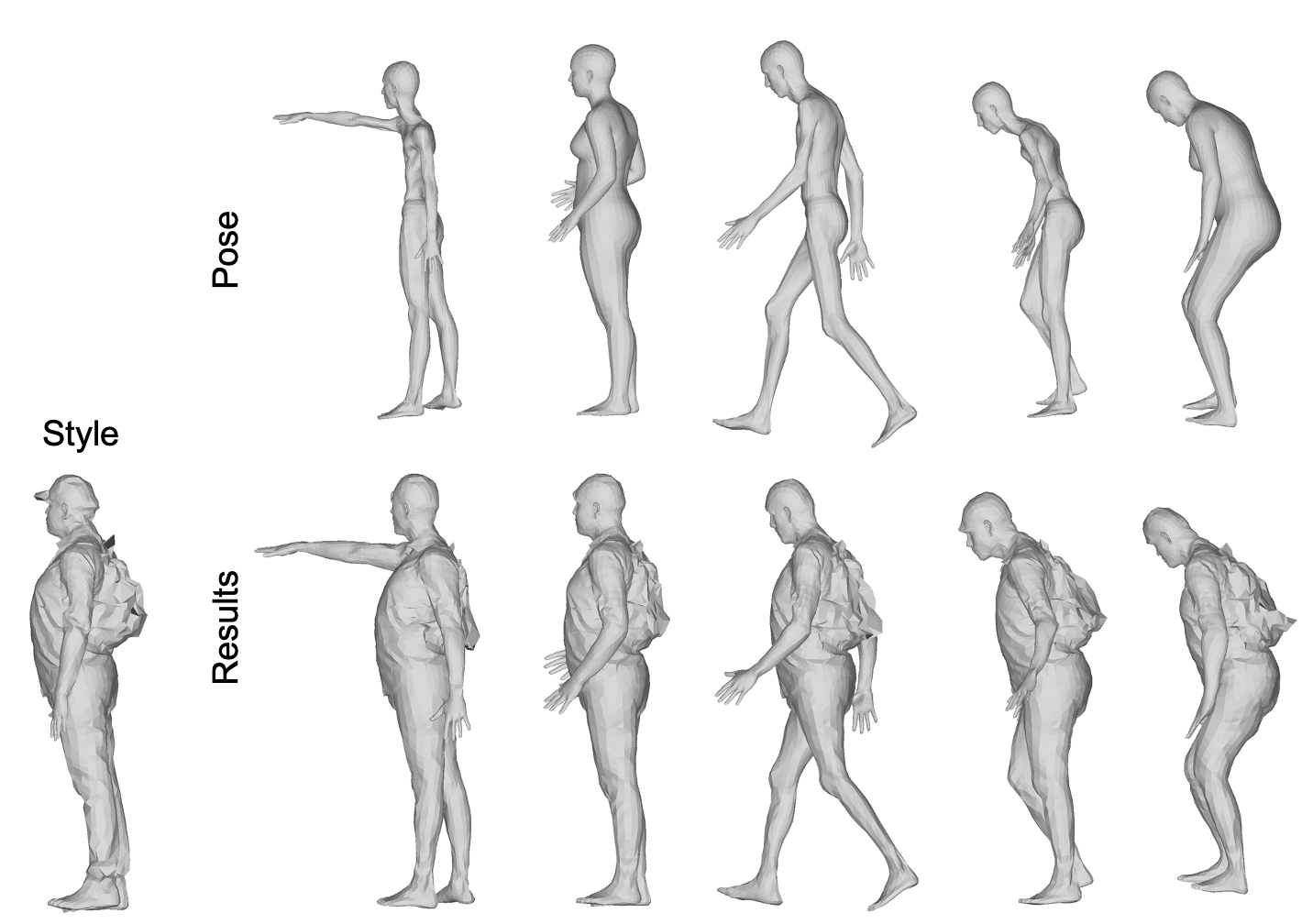}
\end{center}
   \caption{Side-view of male shape, clothing and accessories style transferred to different poses with varied body shapes.}
\label{fig:clothingval2}
\end{figure}

\begin{figure}[H]
\begin{center}
    \includegraphics[width=0.80\linewidth]{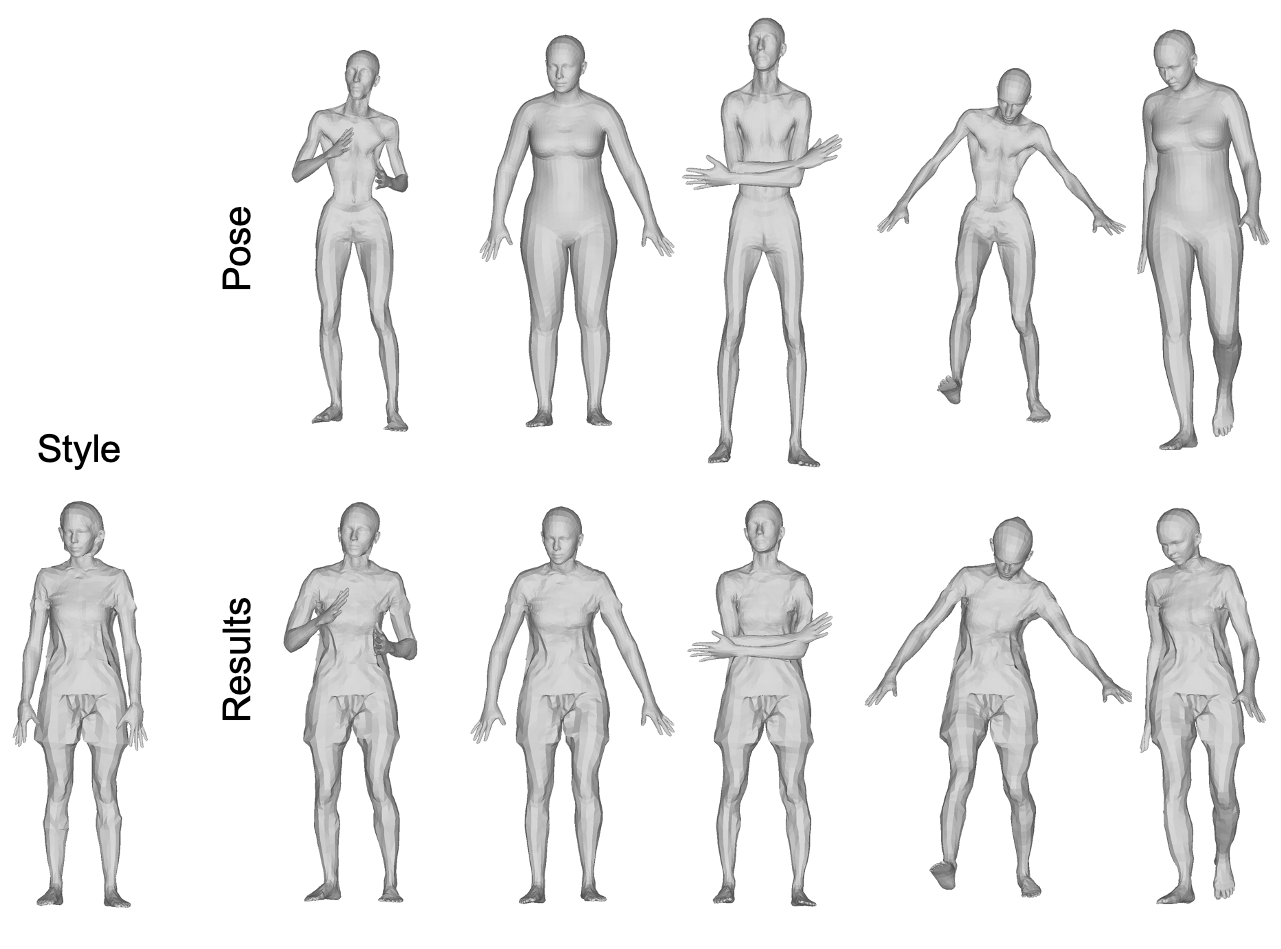}
\end{center}
   \caption{Female shape and clothing style transferred to different poses with varied body shapes.}
\label{fig:clothingval5}
\end{figure}

We qualitative evaluate the proposed network results to the ground-truth samples, see Figures \ref{fig:test1}, \ref{fig:test2}, \ref{fig:clothingval2}, \ref{fig:clothingval5} and  \ref{fig:eval_compare_results}.
The proposed network is able to transfer the shape style of unseen realistic characters from realistic body shapes datasets (FAUST and Dynamic FAUST) \cite{Bogo:CVPR:2014,dfaust:CVPR:2017} and preserve the high-frequency details such as body muscular anatomy in Figures \ref{fig:test1} and \ref{fig:test2}.
It is visible from these figures that the shape style is accurately transferred and consistent across multiple poses with very distinct body shapes.
To evaluate the generalization ability we validate our method in more challenging scenarios, such as realistic clothed people (3DPW) \cite{vonMarcard2018}.
Figures \ref{fig:clothingval2} and \ref{fig:clothingval5} clearly illustrate the ability to transfer realistic shape, clothing and accessories from clothed people \cite{vonMarcard2018}.
Note that at training the network has no information of clothed shapes, hence the results on clothed shapes illustrate the generalization ability of the proposed network.
Figure \ref{fig:clothingval2}, illustrates the ability to represent strong deformation, such as the back-pack on the identity shape, the network clearly maintains the pose and transfers most of the clothing and accessories.
Figure \ref{fig:clothingval5} strengthen the generalization ability where the identity clothing is accurately transferred to all poses.
The proposed method is suitable for motion retargeting, allowing to perform shape style transfer on long animation sequences.
The results are consistent across the animation and do not introduce obvious artefacts, such as foot scatting, jitter or streching.
Please consult the supplementary material for more examples of unseen realistic characters and qualitative results on animation sequences.
\subsection{Shape Deformation Transfer Comparison}
In this section, we compare the proposed method with state-of-the-art methods on shape transfer, such as NPT \cite{Wang_2020_CVPR}, USPD \cite{keyang20unsupervised} and CPST \cite{Basset2019}.
NPT \cite{Wang_2020_CVPR} is re-trained using the AMASS dataset \cite{AMASS:ICCV:2019} generated for this work, and USPD \cite{keyang20unsupervised} is trained on the original AMASS dataset, given that the pre-trained models have a limited range of motion and shapes.
Figure \ref{fig:eval_compare_results} illustrate qualitative comparison results against state-of-the-art approaches \cite{Wang_2020_CVPR,keyang20unsupervised,Basset2019} on shape details transfer. 
It is visible that the proposed results accurately represent the pose and identity shape compared to the ground-truth, whereas the methods proposed by NPT \cite{Wang_2020_CVPR} and USPD \cite{keyang20unsupervised}, fail to preserve the shape detail.
Although the pose representation is arguable comparable, the shape transferred presents strong artefacts, such as, mesh collapse and stretching, which are commonly found issues with traditional linear blend skinning methods.
Therefore, the proposed method being preferred for learning non-linear deformations from the identity shape to transfer to the target pose, allowing realistic human shape deformation and representation.
\section{Limitations}
The method is not capable to avoid intra-body collisions, given that this issue is already present in the dataset, see supplementary material for an illustration.
The dataset generation process does not take into account the sublet differences in the pose for different body morphologies.
Further work would require a dataset with realistic body poses and shapes, and a network to handle this particular problem.
\section{Conclusion}
This work presented a shape style transfer network for the 3D human shape domain.
We demonstrate the ability to transfer realistic human bodies onto posed 3D shapes by learning from a synthetic template with several distinct shapes and augmented with motion capture data.
The proposed approach borrows concepts from traditional image style transfer, such as AdaIN and SPADE architectures, that have proven successful in the image domain.
This work presents a novel network architecture combined with a robust set of losses and a discriminator model that permits learning how to deform a posed shape to match the style of a given identity shape.
Hence, avoiding stretching artefacts from deforming the identity shape to match a given pose, shown in the comparison with state-of-the-art learning-based methods.
The future for this work will focus on pose generalisation to allow the synthesis of extreme poses with a limited amount of training data.

{\small
\bibliographystyle{plain}
\bibliography{arxiv_final}
}

\end{document}